%% file: IR_zhiyao.tex
\documentclass[sigconf,10pt]{acmart}

\hypersetup{draft}
\usepackage{booktabs} 
\usepackage{hyperref}
\usepackage{lipsum}
\usepackage{graphicx}
\usepackage{siunitx, amsmath}
\usepackage{subfigure}
\usepackage{pifont}
\usepackage[normalem]{ulem}
\usepackage{multirow}
\usepackage{multicol}
\usepackage{comment}
\usepackage{color}
\usepackage{etoolbox}
\newbool{inccomment}
\setbool{inccomment}{true}
\newcommand{\hl}[1]{\ifbool{inccomment}{{\color{magenta}#1}}{}}
\newcommand{\fc}[1]{\ifbool{inccomment}{{\color{blue}#1}}{}}
\usepackage{bbm}

\usepackage{tabularx}
\usepackage{graphicx}
\usepackage{supertabular}

\usepackage{algorithm}
\usepackage{algpseudocode}
\algtext*{EndWhile}
\algtext*{EndIf}
\algtext*{EndFor}
\algtext*{EndProcedure}
\usepackage[super]{nth}
\usepackage{longtable}

\newcommand{\thistheoremname}{}
\newtheorem*{genericthm*}{\thistheoremname}
\newenvironment{namedthm*}[1]
  {\renewcommand{\thistheoremname}{#1}%
   \begin{genericthm*}}
  {\end{genericthm*}}
 
\usepackage{mathtools}

\graphicspath{{./_fig/}}

\usepackage{amsthm}

\allowdisplaybreaks
\makeatletter
\def\BState{\State\hskip-\ALG@thistlm}
\makeatother

\copyrightyear{2020} 
\acmYear{2020} 
\setcopyright{acmcopyright}\acmConference[ICCAD '20]{IEEE/ACM International Conference on Computer-Aided Design}{November 2--5, 2020}{Virtual Event, USA}
\acmBooktitle{IEEE/ACM International Conference on Computer-Aided Design (ICCAD '20), November 2--5, 2020, Virtual Event, USA}
\acmPrice{15.00}
\acmDOI{10.1145/3400302.3415763}
\acmISBN{978-1-4503-8026-3/20/11}

\title{Fast IR Drop Estimation with Machine Learning}
\subtitle{Invited Paper \vspace{-3mm}}

\author{ Zhiyao Xie$^\dagger$, Hai Li$^\dagger$, Xiaoqing Xu$^\ddagger$, Jiang Hu$^\mathsection$, Yiran Chen$^\dagger$}

\affiliation{\vspace{2mm}
\institution{$^\dagger$ Duke University, Durham, NC, USA}
\vspace{0.5mm}
\institution{$^\ddagger$ ARM Inc., Austin, TX, USA}
\vspace{0.5mm}
\institution{$^\mathsection$ Texas A\&M University, College Station, TX, USA}
\vspace{1mm}
\institution{zhiyao.xie@duke.edu  \; \; \;  hai.li@duke.edu}
}

\begin{document}


\input{_txt/abstract}
\maketitle

\input{_txt/0_introduction}

\input{_txt/2_method}

\input{_txt/4_dynamic}

\input{_txt/5_future_works}

\input{_txt/6_conclusion}



\bibliographystyle{ACM-Reference-Format}

\bibliography{IR_zhiyao.bib}
\end{document}

%% file: _txt/abstract.tex
\begin{abstract}

IR drop constraint is a fundamental requirement enforced in almost all chip designs. However, its evaluation takes a long time, and mitigation techniques for fixing violations may require numerous iterations. As such, fast and accurate IR drop prediction becomes critical for reducing design turnaround time. 
Recently, machine learning (ML) techniques have been actively studied for fast IR drop estimation due to their promise and success in many fields.
These studies target at various design stages with different emphasis, and accordingly, different ML algorithms are adopted and customized. This paper provides a review to the latest progress in ML-based IR drop estimation techniques. It also serves as a vehicle for discussing 
some general challenges faced by ML applications in electronics design automation (EDA), and demonstrating how to integrate ML models with conventional techniques for the better efficiency of EDA tools.

\end{abstract}

%% file: _txt/0_introduction.tex
\section{Introduction}

IR drop, or voltage drop, is the deviation of a power supply level from its specification that occurs when current flows through power grids.  It must be restricted in order for a circuit to meet its timing target and function properly.
As design and manufacturing technologies advance, the increased current load further exaggerates IR drop violations, which become a critical concern for both VLSI design and test~\cite{tehranipoor2010power}.

In order to meet IR drop constraints, designers need to estimate and mitigate IR drop throughout design stages from placement to signoff in multiple iterations. It may also be measured during post-silicon verification. Obtaining an accurate estimation of IR drop through simulation-based commercial tools is very time consuming~\cite{Redhawk, fang2018machine, xie2020powernet}. Thus, IR drop mitigation guided by frequent IR drop simulations is computationally costly and hampers the overall design turnaround time. To speed up this process, a fast yet accurate IR drop estimator becomes a critical need.

In recent years, machine learning (ML) applications in electronics design automation (EDA) have started to attract wide attention. They have enabled fast estimation on many important metrics for chip design, including timing~\cite{barboza2019machine}, power~\cite{zhou2019primal, kim2019simmani}, design rule violation~\cite{xie2018routenet, yu2019painting, liang2020drc}, crosstalk~\cite{liang2020routing}, testability~\cite{ma2019high}, lithography hotspots~\cite{yang2017imbalance, zhang2016enabling}, clock tree's quality~\cite{lu2019gan}, placement solution~\cite{mirhoseini2020chip}, routing solution~\cite{zhu2019geniusroute}, and IR drop~\cite{ho2019incpird, pao2020xgbir, yamato2012fast, dhotre2017identification, lin2018ir, fang2018machine, mozaffari2019efficient, xie2020powernet}. There have been many ML-based IR drop estimators targeting at various design stages with different emphasis. The majorities claim orders-of-magnitude acceleration compared with simulations-based solutions provided by widely-adopted commercial tools \cite{Redhawk}.

\begin{figure}[!b]
  \centering
    \includegraphics[width=0.9 \columnwidth]{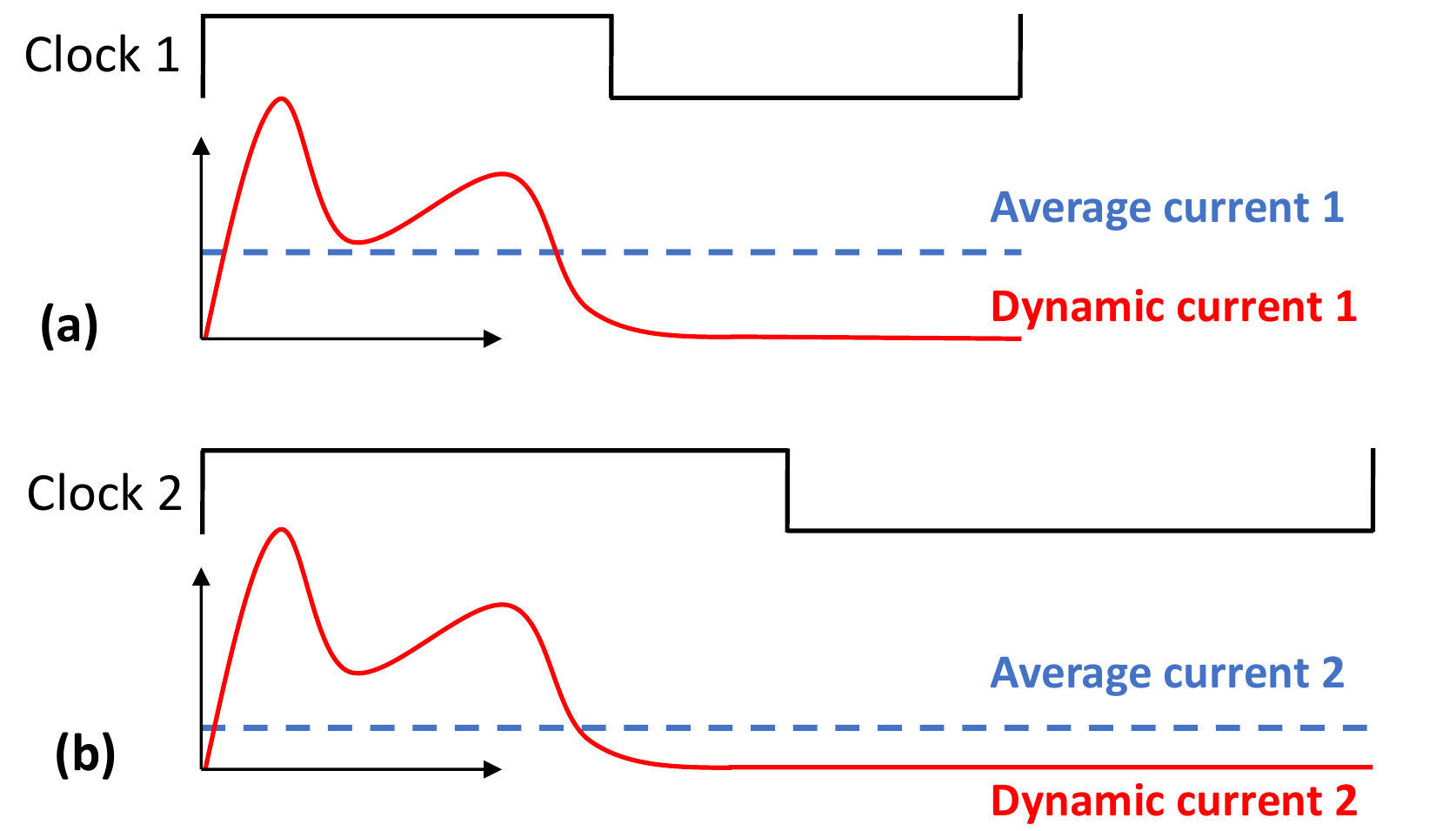}
  \captionsetup{justification=centering}
  \caption{Static and dynamic analysis on current~\cite{nithin2010dynamic}.}
  \label{static}
\end{figure}

\begin{table*}[t]
\caption{Comparison Among Different Works on IR drop estimation}
\label{tbl:summary}
\centering
\begin{tabular}{| p{0.16\linewidth} |  p{0.13\linewidth} | p{0.04\linewidth} | p{0.15\linewidth} | p{0.07 \linewidth}
                | p{0.15\linewidth}  |  p{0.14\linewidth} |}
\hline
\multirow{2}{*}{Methods}  &  \multirow{2}{*}{Type of IR Drop}  &  \multirow{2}{*}{Time}  &  \multirow{2}{*}{ML Model}  & Cross- &   \multirow{2}{*}{Features} &  \multirow{2}{*}{Objective} \\
                 &            &         &     &   Design   &   & \\
\hline
IncPIRD \cite{ho2019incpird}                          & \multirow{2}{*}{Static IR}       & 2019    
       &  XGBoost          & Yes  &  $I_g$, $R_g$, PDN, $G$   & IR mitigation   \\
XGBIR \cite{pao2020xgbir}                             &                                  & 2020     
       &  XGBoost          & -    &  $I_g$, $R_g$, PDN        & PDN design   \\
\hline
Yamato \emph{et al.} \cite{yamato2012fast}            &  \multirow{4}{*}{Vector-based IR} & 2012
       &  Linear Regression &  No  & $P_c$  & IR-aware timing  \\
    
Dhotre \emph{et al.} \cite{dhotre2017identification}  &                                   & 2017 
       &  Clustering       & Yes  & $r_{tog}$, $c$  & IR prediction  \\
Lin \emph{et al.}  \cite{lin2018ir}                   &                                   & 2018
       &  ANN              & No   & $P_c$, $R_c$, $t_c$, $r_{tog}$, $c$ & IR mitigation  \\
Fang \emph{et al.} \cite{fang2018machine}             &                                   & 2018     
       &  CNN, XGBoost     & No   & $P_c$, $I_c$, $R_c$, $t_c$, $r_{tog}$, $c$ & IR mitigation \\
\hline
PowerNet \cite{xie2020powernet}                       &  Vectorless IR                   & 2020     
       &  CNN              & Yes  &  $P_c$, $t_c$, $r_{tog}$  & IR mitigation \\
\hline
Mozaffari \emph{et al.} \cite{mozaffari2019efficient} &  Sillicon PSN                    & 2019     
       &  ANN, CNN, NLP    & Yes  &  $r_{tog}$, $G$           & PSN prediction \\
\hline
\end{tabular}
\end{table*}

The ML-based IR drop estimators can be classified into two major categories, based on whether they estimate static IR drop or dynamic IR drop.
Figure \ref{static} shows a comparison between the average current and dynamic current under different frequencies~\cite{nithin2010dynamic}. The static IR drop analyses in most commercial tools only measure the average current drawn from power grids without considering switching activities~\cite{Cadence, nithin2010dynamic, Redhawk}. It is widely used to identify the weakness of a power delivery network (PDN) at an early design stage when switching vectors are not available~\cite{Cadence, Redhawk}. There have been many traditional methods for fast power grids analysis \cite{zhao2002hierarchical, qian2005power, chen2001efficient, zhuo2008power, su2003power, kozhaya2002multigrid}. In contrast, dynamic IR drop captures the peak transient current value based on switching activities. Thus, it is a more strict constraint and more difficult to predict \cite{lin2004full, nithin2010dynamic}. The significant difference between static and dynamic IR drop leads to distinctive problem settings and corresponding ML solutions. We will introduce methods on static IR drop and dynamic IR drop separately.

Power supply noise (PSN) is sometimes also loosely referred to as IR drop~\cite{mozaffari2019efficient}. But the PSN actually comprises both $L* di/dt$ and the dynamic IR drop \cite{chen1997power}. The $L* di/dt$ component
is an inductive effect caused by rapid current changes through power grids. Most works reviewed in this paper focus on the IR drop without considering $L* di/dt$.

In this paper, we will summarize the latest progress in ML-based techniques for both static and dynamic IR drop. The remainder of this paper is organized as follows. Section 2 presents an overview of the ML-based IR drop estimators. Section 3 covers these estimators in detail. In Section 4, we discuss some challenges and the integration of ML estimators into EDA tools.

%% file: _txt/2_method.tex
\section{Method Overview}

\label{sec:method}

\subsection{Estimation on IR Drop}

Table \ref{tbl:summary} summarizes the ML-based IR drop estimators from different perspectives. The criteria shown in Table \ref{tbl:summary} are introduced below.

\subsubsection{Type of IR Drop} 

Different types of IR drop will result in largely different problem settings and ML solutions. Table \ref{tbl:summary} covers both the static and dynamic IR drop. For dynamic IR drop, the commercial simulators \cite{Redhawk, Cadence}  provide both vector-based and vectorless options. The vector-based analysis relies on switching vectors as input, which is limited by the late availability of switching activities in many design processes \cite{xie2020powernet, Redhawk, Cadence}. The vectorless option allows simulation by the toggling probability instead of switching vectors \cite{Cadence, lin2004full}. All estimators on both static and dynamic IR drop use the simulation result from the commercial tool as the ground-truth label for training. One exception is Mozaffari \emph{et al.} \cite{mozaffari2019efficient}, which directly uses the PSN measured on silicon as their label. This method is also included in the comparison among dynamic IR drop estimators.

\subsubsection{ML Model} 
These methods can be differentiated by their ML models. The frequently used ML algorithms include artificial neural network (ANN), XGBoost \cite{chen2016xgboost}, and convolutional neural network (CNN) \cite{krizhevsky2012imagenet}. ANN and XGBoost typically process one-dimensional inputs, while CNN handles two-dimensional features with spatial information. In addition, the NLP model in Table \ref{tbl:summary} means neural network models used in natural language processing.

Here we briefly introduce these frequently applied ML models. ANN is also referred to as multilayer perceptron (MLP). It consists of multiple layers of nonlinearly-activating nodes. XGBoost is an efficient implementation of gradient boosting decision tree (GDBT) \cite{friedman2001greedy}, which builds decision trees sequentially, with each tree built on the error of the previous ones. A typical structure of CNN is composed of convolutional layers, pooling layers, and fully-connected (FC) layers. Convolutional and pooling layers perform downsampling, and FC layers at the end generate the output.

\subsubsection{Cross-Design} 

A cross-design ML estimator means the estimator applies to new designs that are not in the training set. Here we require the \emph{new} design to be different from training designs at the netlist level. Thus the following examples are not viewed as cross-design: 1) Models trained and tested with different layout implementation of the same netlist. 2) Models trained and tested on multiple designs, which appear in both training and testing set.

\subsubsection{Feature Selection} 

Features selection is one of the most critical steps in ML applications, which decides models' performance and efficiency. Table \ref{tbl:summary} provides a high-level summary of the feature types extracted in each method. Some straightforward features, like the location of cells, are not listed. These features are introduced below.

\begin{itemize}
\item \emph{Resistance} \{$R_c$, $R_g$\} \emph{-} Resistance is a determining factor of IR drop. Some works directly use the total resistance measured on the path from power pad to each cell instance as input, denoted as $R_c$. Some others use different types of resistance measured on power grids and power nodes, denoted as $R_g$.
\item \emph{Current} \{$I_c$, $I_g$\} \emph{-} Current $I$ is the other determining factor for IR drop. Like resistance, some works use the average or peak current measured on each cell instance as input, denoted as $I_c$. Some others use current loads or the total current on power grids, referred to as $I_g$.
\item \emph{Power} \{$P_c$\} \emph{-} Power dissipation is an important input in IR drop simulation. It correlates with current and is easier to simulate. The power consumption on cell instance is $P_c$. Power dissipation may include internal power, switching power, and leakage power.
\item \emph{Time} \{$t_c$\} \emph{-} The timing interval or window of switching for each cell instance. It comprises of estimated earliest and latest signal arrival time.
\item \emph{PDN} - The information about PDN. It includes geometric information about power grids, voltage source, and current loads. It also includes metal and via resistance on the PDN.
\item \emph{Toggling activity} \{$r_{tog}$\} \emph{-} The switching activity of each cell instance. It is usually measured by the toggle rate, the number of toggles of each cell divided by the number of cycles within a timing window. It can also represent the percentage of toggled flip-flops versus overall flip-flops. 
\item \emph{Cell information} \{$c$\} \emph{-} The information about each cell instance except current, power and resistance. It includes area, cell load, cell type, etc.
\item \emph{Global information} \{$G$\} \emph{-} The information about the whole design or layout. It includes the process, voltage, temperature, frequency, and the size of the layout.
\end{itemize}

\subsubsection{Objective}

Besides estimating different types of IR drop with various input features, previous works also assume different application scenarios for their methods. Below are their objectives with their IR drop estimators. 

\begin{itemize}
\item \emph{Prediction -} Some works focus on estimation and do not give details on how the estimation will be applied.
\item \emph{Mitigation -} Most works estimate IR drop hotspots to guide mitigation. Mitigation solutions include power grid enhancement, cell or macro movement, decoupling cap insertion, etc.
\item \emph{PDN Design -} Some works focus on optimizing PDN structure with the fast IR drop estimation. This is similar to IR drop mitigation by power grid enhancing, but it usually takes multiple different PDN structures as candidates for evaluation.
\item \emph{Timing -} Some works estimate IR drop or PSN so that they can measure timing based on a more accurate supply voltage for each instance.
\end{itemize}

\subsection{IR Drop-Related Estimation}

\begin{table}[!tb]
  \centering
  \caption{IR Drop-Related ML Works}
  \label{tbl:related}
  \resizebox{\linewidth}{!}{
  \begin{tabular}{  l  | l | l  }
 	\hline
 	Works   & ML Model    & Label to estimate   \\
\hline
\multirow{2}{*}{Lee \emph{et al.} \cite{lee2020learning}} &  Special & Weights under  \\
                                         & Neural Network  &  distortion \\
	\hline
Chan \emph{et al.} \cite{chan2016learning}    & Boosting SVM  & \multirow{3}{*}{IR-aware timing}     \\ 
Liu \emph{et al.} \cite{liu2016psn}           &  ANN, SVM  &      \\  
Ye \emph{et al.} \cite{ye2014chip}            &  SVM      &      \\
	\hline
\multirow{2}{*}{Chang  \emph{et al.} \cite{chang2017generating}} &  Gaussian process    &    Routing      \\
                  &          regression    &  wirelength \\
\hline
\multirow{2}{*}{Chhabria \emph{et al.} \cite{chhabria2020template}}  &  \multirow{2}{*}{CNN}  & Optimal PDN \\
                      &      & Template   \\
\hline
  \end{tabular}
  }
\end{table}

Besides the IR drop estimators covered in Table \ref{tbl:summary}, some other ML works consider IR drop without directly estimating it. Instead, they predict some design objectives related to IR-drop. A summary of IR drop-related ML works is provided in Table \ref{tbl:related}.

\subsubsection{Impact of IR Drop}

Instead of predicting IR drop, some works directly estimate the impact of it. The work in \cite{lee2020learning} predicts the distorted multiplication caused by the IR drop on ReRAM-based neural networks. Some other works directly estimate timing under the impact of IR drop and/or Ldi/dt, which is referred to as PSN-aware timing analysis or IR-aware timing analysis~\cite{chan2016learning, liu2016psn, ye2014chip}. Notice the difference between \cite{chan2016learning, liu2016psn, ye2014chip} and \cite{yamato2012fast}. The method in \cite{yamato2012fast} estimates dynamic IR drop and uses the predicted IR drop to estimate timing, while models in \cite{chan2016learning, liu2016psn, ye2014chip} directly estimate IR drop-affected timing. They use completely different labels during training.

\subsubsection{IR Drop as A Constraint}

Some other works only view IR drop as one constraint and focus on other design objectives. For example, when designing PDN with machine learning models, works in \cite{chang2017generating, chhabria2020template} optimize routing resources and PDN qualities while considering the IR drop and electromigration (EM) as constraints. As shown in Table \ref{tbl:related}, what they actually predict is the routing wirelength \cite{chang2017generating} or the optimal PDN template \cite{chhabria2020template}. After that, their solutions are verified to meet IR drop specifications.

%% file: _txt/4_dynamic.tex
\begin{figure*}[tb]
  \centering
    \includegraphics[width=\textwidth]{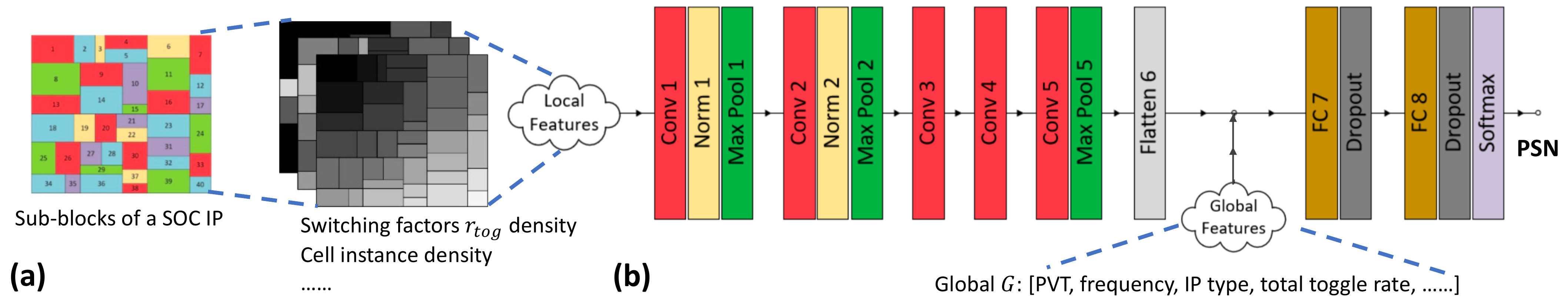}
 \captionsetup{justification=centering}
  \caption{Mozaffari \emph{et al.} (a) Generate density maps based on sub-blocks. \\ (b) The CNN method with both local and global features. \cite{mozaffari2019efficient}}
  \label{mozaffa}
\end{figure*}

\section{Estimators in Detail}

\subsection{Static IR Drop Estimation}

The two representative methods \textbf{IncPIRD}~\cite{ho2019incpird} and \textbf{XGBIR}~\cite{pao2020xgbir} on static IR drop perform fast power grid analysis with ML models. Unlike works on dynamic IR drops, their features and labels are measured on the power nodes on PDN instead of on cells.

These two works \cite{ho2019incpird, pao2020xgbir} share many similarities in terms of both features selection and the ML model. Table \ref{tbl:feature1} and \ref{tbl:feature2} describe the major features used in each work, respectively. They both use two major types of features: one is the PDN's topology, the other is electrical features. For electrical features, they both define features representing the \emph{pullup} and \emph{pulldown} for each analyzed node, based on the amount of current drawn and the effective resistance between each power node, voltage sources, and current loads. Lastly, they both adopt the XGBoost model.

\begin{table}[!b]
  \centering
  \caption{Description of Features of IncPIRD \cite{ho2019incpird}}
  \label{tbl:feature1}
  \vspace{-2mm}
 \resizebox{0.95\linewidth}{!}{
  \begin{tabular}{  l  | l   }
  \hline
 	Category      &   Description of Feature on node $n$        \\
	\hline
\multirow{2}{*}{Chip / PDN} &  Pitch of all metal layers    \\
                            &  Width / height of the chip                 \\
\hline
\multirow{3}{*}{Electrical} & \emph{Pullup}: The effective resistance at $n$  \\
                            & \emph{Pulldown}: The symbolic IR drop at $n$   \\
                            & Pullup and Pulldown of $n$'s neighbors \\
\hline
\multicolumn{2}{c}{}\\
  \end{tabular}
  }
\end{table}

\begin{table}[!b]
    \vspace{-5mm}
  \centering
  \caption{Description of Features of XGBIR \cite{pao2020xgbir}}
  \label{tbl:feature2}
    \vspace{-2mm}
\resizebox{0.95\linewidth}{!}{
  \begin{tabular}{  l  | l   }
 	\hline
 	Category      &   Description of Feature on node $n$        \\
	\hline
\multirow{3}{*}{Chip / PDN} &  The number of power tracks        \\
                            &  Distance between $n$ and boundary  \\
                            &  Power track segmental resistance    \\
\hline
\multirow{4}{*}{Electrical} & \emph{Pullup}: Voltage sources' impact on $n$  \\
                            & \emph{Pulldown}: Current loads' impact on $n$   \\
                            & \emph{V2I}: Resistance  between voltage   \\
                            & \quad\quad sources and current loads \\
\hline
  \end{tabular}
  }
\end{table}

Based on its XGBoost model, \textbf{IncPIRD} \cite{ho2019incpird} proposes a flow to mitigate IR drop incrementally with many iterations. In each iteration, based on the prediction, the designer performs incremental mitigation and generates an updated layout. To avoid retraining the ML model in each iteration, it defines a model update condition, determining whether the existing ML model still applies to the updated layout. The model is only retrained when testing data satisfies the update condition. Such an update condition actually considers the model's robustness quantitatively, which is rarely discussed in most ML applications in EDA.

\subsection{Dynamic IR on Individual Cells}

\begin{figure*}[!h]
  \centering
    \includegraphics[width=0.95\textwidth]{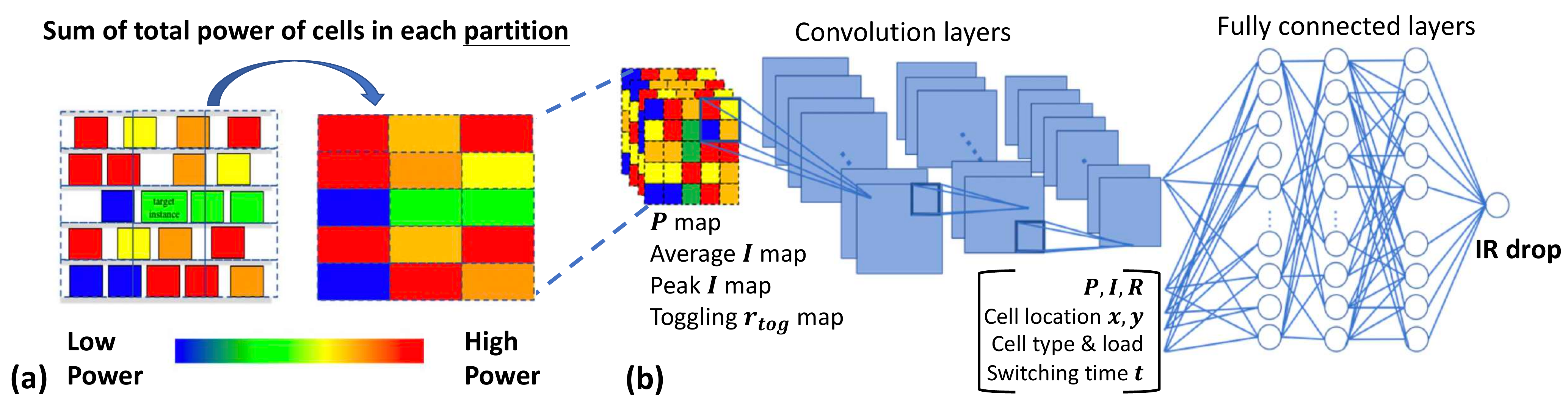}
  \caption{Fang \emph{et al.}  (a) Generate power maps around each cell. (b) ML model for each cell. \cite{fang2018machine}}
  \label{fang}
\end{figure*}

\textbf{Yamato \emph{et al.}} \cite{yamato2012fast} provide an intuitive method for dynamic IR drop estimation. According to their observation, most cells show a high correlation between their power and IR drop. It indicates that a simple linear model on power for each individual cell will perform well. Assuming $m$ is the number of test patterns for training, each cell's training set consists of $m$ power numbers as input and $m$ corresponding IR drop labels. Then a linear model is trained for it.

Considering the huge number of cell instances in industrial designs, devoting a separate model for each cell can be costly in computation. Designers may not need such a fine granularity in practice. But the simplicity of linear models compensates the computation cost to a certain extent. 
In addition, one unique model for each cell instance may not generalize very well. If there is a small change in the placement or the PDN, the power-IR relations of some affected cells change accordingly; then the relevant models may need to be retrained.

The model for each individual cell ignores the impact of its neighboring cells. The IR drop at each cell instance depends on the current demand in a region enclosing the cell. For each region, all cells inside contribute to the overall current demand, which flows through a common metal path on PDN. Thus, all cells in a high IR drop region suffer from an excessive IR drop. To capture this, many works consider the IR drop in the granularity of regions or partitions instead of individual cells.


\subsection{Dynamic IR on Regions}

\textbf{Dhotre \emph{et al.}}~\cite{dhotre2017identification} assign all cell instances to multiple clusters by clustering algorithms like k-means or DBSCAN \cite{ester1996density}. Each cell's power is approximated by its toggling activity. All cells are clustered based on their locations, area, and toggling activity. Then the regions with high-toggling-activity clusters are viewed as both power and IR drop hotspots. This method considers locality information by clustering the neighboring cells with similar information. As an unsupervised algorithm, it applies to different designs.

\textbf{Mozaffari \emph{et al.}} \cite{mozaffari2019efficient} adopt more complex models. Figure \ref{mozaffa} shows their method with CNN structure. For each test pattern, the input includes the information from the whole layout, and accordingly, the label is the PSN value measured on the chip. Hence, this method is very coarse-grained. In Figure \ref{mozaffa}(a), it builds two-dimensional `local features' with detailed information from each sub-block. The features include togging rate density and cell density. It also builds a vector with the chip-level global information and total toggling rate, representing `global features'. In Figure \ref{mozaffa}(b), the two-dimensional `local features' firstly go through convolutional layers, then the flattened output is concatenated with one-dimensional `global features' for the fully-connected layers. In this way, the CNN model learns from features in different dimensions, from both local and global perspectives. By incorporating global features of the whole chip, it claims that this method applies to new chips, which means the method is cross-design.

\textbf{Dhotre \emph{et al.}}~\cite{dhotre2017identification} and \textbf{Mozaffari \emph{et al.}} \cite{mozaffari2019efficient} consider the localized power demand by estimating IR drop in regions or partitions. However, as introduced in Figure \ref{static}, the dynamic IR drop captures the worst transient IR drop. 
The high-current-consumption cells in the same region may not switch simultaneously, which does not result in a transient peak in IR drop. It means models may give false alarms on regions with high power consumption if they do not consider the signal arrival time of cells.

\subsection{Dynamic IR with Timing}

\textbf{Fang \emph{et al.}} \cite{fang2018machine} estimate the IR drop of each cell by considering both neighboring cells' current consumption and the switching timing window of cells. The timing window of each cell includes the minimum and maximum signal arrival time. Figure \ref{fang}(a) shows how it generates feature maps for each cell. The entire layout is tessellated into an array of small partitions. Then the values of power, peak current, average current, and toggling rate within each partition are calculated. 
For each cell, to quantify the impact from neighboring cells in the local region, features maps are constructed based on the $21\times11$ local partitions around this cell. These maps are the input of the CNN model. 
In addition, all the available information about the estimated cell is stacked into a vector, including its power, current, resistance, switching time, location, etc. This vector is provided through FC layers, as shown in Figure \ref{fang}(b).

When comparing Figure \ref{fang} to Figure \ref{mozaffa}, \textbf{Mozaffari \emph{et al.}} \cite{mozaffari2019efficient} and \textbf{Fang \emph{et al.}} \cite{fang2018machine} use highly similar CNN architectures to incorporate both two-dimensional and one-dimensi\allowbreak onal inputs. The two-dimensional inputs in both works describe the distribution of power or toggling activity. The difference is that one of them estimates the IR drop on each cell, while the other estimates that on the entire layout. Due to such difference, the vector provided to FC layers in Figure \ref{fang}(b) describes each cell, while in Figure \ref{mozaffa}(b) it provides global features about the entire chip.

One significant difference between \textbf{Fang \emph{et al.}}~\cite{fang2018machine} and the aforementioned methods is the employment of instances' switching time window. It enables this model to capture simultaneous switchings among cells. However, the switching time and cell locations themselves do not directly correlate with IR drop. Providing them directly to FC layers may overfit the ML model to the training data, which potentially limits the generalization of the model. Instead of being cross-design, this model infers the same design in training, with merely cell location changes.



\begin{figure*}[!h]
  \centering
    \includegraphics[width=0.95\textwidth]{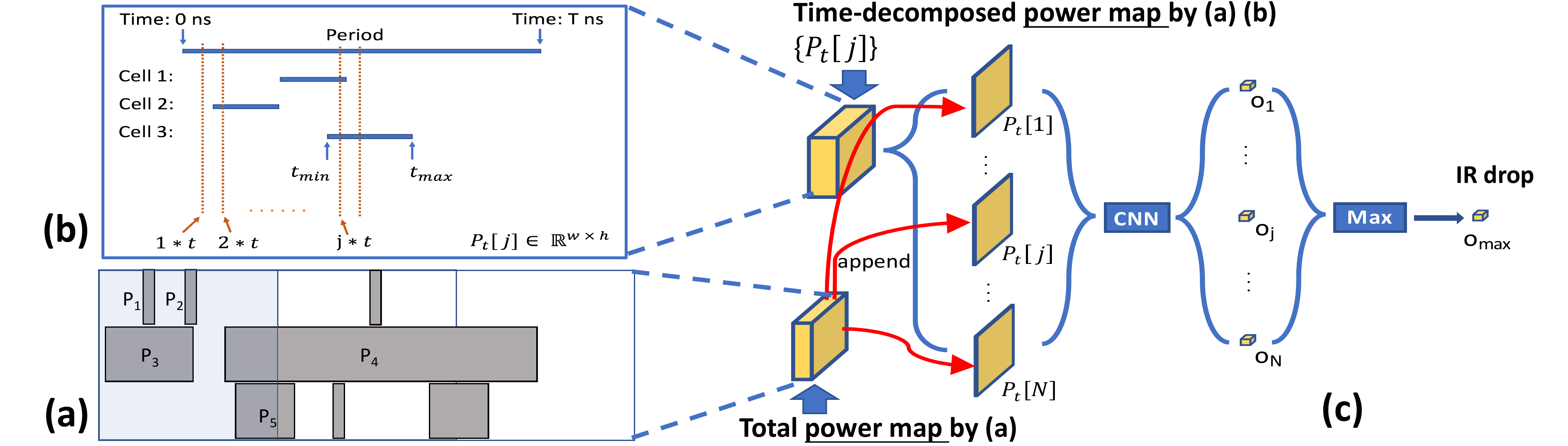}
  \captionsetup{justification=centering}
  \caption{PowerNet method.  (a) Generate power maps around each grid. \\ (b) Timing-decomposed power map. (c) ML model for each grid. \cite{xie2020powernet}}
  \label{powernet}
\end{figure*}

\subsection{Cross-Design Dynamic IR with Timing}

\textbf{PowerNet} \cite{xie2020powernet} chooses a different direction to utilize timing information. Instead of directly using switching time as the input, it incorporates the switching time of cells into multiple power distribution maps, or named power maps.

In Figure \ref{powernet}(a), the layout is firstly tessellated into an array of uniform grid tiles, then the internal, switching, and leakage power are measured for each tile. For each tile, its input features include the power of the $30\times30$ local tiles around it, denoted as `total power map' in Figure \ref{powernet}(c). This process is similar to the feature generation scheme in \textbf{Fang \emph{et al.}} \cite{fang2018machine}. Besides that, power maps with timing information are constructed in Figure \ref{powernet}(b). It divides one clock cycle into $N$ different time instants and measures $N$ corresponding time-decomposed power maps $P_t[1]-P_t[N]$. For each instant $j$, only the cells that can possibly switch at that instant contribute their power consumption to $P_t[j]$. Thus, the power map $P_t[j]$ only measures the dynamic power consumed at the $j$'s instant.

Based on the time-decomposed power maps $P_t[1]-P_t[N]$ corresponding to $N$ instants, \textbf{PowerNet} \cite{xie2020powernet} captures the one that leads to the highest transient IR drop at each grid. To achieve this, the same CNN model processes all $N$ time-decomposed power maps in parallel, as shown in Figure \ref{powernet}(c), each output corresponds to the transient IR drop at that instant. Then the maximum among $N$ outputs is the final estimation. Notice the `total power map' without timing information is also used as input together with every time-decomposed power map $P_t[j]$ for each instant $j$.

By devoting a time-decomposed power map for each instant, \textbf{PowerNet} \cite{xie2020powernet} uses a CNN model with the maximum structure to capture both localized power demand and simultaneous switching activities. Both of them lead to dynamic IR drop hotspots. But as a trade-off, compared with a regular CNN model, processing $N$ power maps in parallel also leads to higher computation cost.

%% file: _txt/5_future_works.tex
\section{Rethinking ML for IR Drop}

\subsection{Challenges and Future Directions}

The ML methods have achieved significant progress in IR drop estimation. Besides further improving the models with more advanced techniques, here we discuss three more general challenges that can be addressed in the future. They may also apply to other ML applications in EDA.

\subsubsection{Evaluation and Comparison} We have discussed many estimators on both static and dynamic IR drops. But it is difficult to make a fair comparison among them and figure out the best solution in different application scenarios. Although there are already many open-sourced benchmarks, researchers synthesize these designs with different design parameters and technology libraries, which result in distinct training and testing data. This can be addressed by an open-source benchmark suite dedicated to ML applications on multiple design objectives. It will enable rapid and clear comparisons among different methods, ensure high-quality training data, relieve researchers from data generation, and attract people from the ML community to contribute.

\subsubsection{Model Development and Maintenance} Most aforementioned IR drop estimators are tuned manually in terms of both feature selection and model architecture on a specific dataset. Later on, when the training dataset is updated, or the application scenario is adjusted, the ML model may need to be fine-tuned or even re-developed from scratch. The development and maintenance of every single ML model heavily rely on human expertise. As ML estimators gain popularity in EDA, the high engineering cost on development and maintenance will grow accordingly. This can be addressed by algorithms like AutoML~\cite{he2019automl}, which automatically search the optimal features combination and ML model structure for any given dataset.

\subsubsection{Model Robustness}  The aforementioned IR drop estimators are verified to be accurate based on their own application scenarios. However, the estimator is not likely to perform well on every test case it sees, especially for those cross-design models. This is because designs differ greatly from each other in terms of architecture, technology node, design flow, etc. Hence, it can be risky every time when the estimator infers a brand new test case. To address this, we consider developing an algorithm to measure how likely the model will perform well on a new test case. This can be achieved by quantifying the similarities between the test case and training data. It bears similarity to the idea of `update condition' in IncPIRD~\cite{ho2019incpird}.

\subsection{Integrating ML Model into EDA Tools}

Most of the IR drop estimators we introduced are standalone studies. Based on existing ML models with good performance, an interesting topic is how to integrate the models with existing EDA techniques to obtain more efficient EDA tools. 

\subsubsection{Early Report} 

For an ML-integrated EDA tool, the most straightforward application is to allow queries on predicted design objectives like IR drop at very early design stages. In addition, EDA tools can enable IR drop-aware timing analysis based on predicted IR drop, as proposed by Yamato \emph{et al.}~\cite{yamato2012fast}.

\subsubsection{Objective-Oriented Design Guidance}

ML models can provide guidance for each individual design stage at a low cost.
For example, IR estimators may enable an IR drop-aware design flow, which allows designers to address IR drop at all necessary design stages, from placement to signoff. In this case, mitigation by cell movement at IR drop hotspots can happen as early as placement, and the power strap enhancement can be done during PDN design in the early iterations. This would largely reduce the number of design iterations compared with addressing IR drop at the signoff stage. Some of these IR-drop-aware methods may already be available in the backend tools, but they may be expensive to use without ML models as their IR drop estimators.

\subsubsection{Design Flow Guidance}

ML models can guide decisions in design flows by their early feedback. ML-integrated tools may allow a fast sweep over multiple design flow settings. In modern industrial chip design, the design flow tuning requires to go through a large amount of parameter setting, and the evaluation of every single trial is costly. ML estimators can evaluate candidate design flows at an early stage, then prune away the flows predicted to be worse than a certain threshold. Such evaluation may require multiple ML models on different design objectives.

\subsubsection{Model Training}

When starting on a new design, the prediction may be based on some cross-design ML methods trained on other similar designs. After a few design iterations, based on collected features and labels, the tool may build a non-cross-design model dedicated to this specific design. As iteration continues, the model may be updated by retraining, or incrementally improved by fine-tuning or ensemble learning.

%% file: _txt/6_conclusion.tex
\section{Conclusion}

In this paper, we summarize the latest progress in developing ML models for fast IR drop estimation. We introduce the innovations and technical details of representative methods in two main categories, static and dynamic IR drop. In addition, we discuss some general challenges in ML estimators and how they may be better integrated into current EDA tools.